\documentclass[letterpaper, 10 pt, conference]{ieeeconf}  

\IEEEoverridecommandlockouts                              

\usepackage{cite}
\usepackage{amsmath,amssymb,amsfonts}
\usepackage{graphicx}
\usepackage{algorithm}
\usepackage{algorithmic}

\begin{document}
\title{
Random Network Distillation Based Deep Reinforcement Learning for AGV Path Planning\\
}

\author{Huilin Yin$^{1}$,
        Shengkai Su$^{1}$,
        Yinjia Lin$^{1}$,
        Pengju Zhen$^{1}$,
        Karin Festl$^{2}$,
        Daniel Watzenig$^{2}$ 
\thanks{$^{1}$ Huilin Yin, Shengkai Su, Yinjia Lin and Pengju Zhen are with the School of Electronic and Information Engineering, Tongji University, Shanghai, China.}
\thanks{$^{2}$ Karin Festl and Daniel Watzenig are with the Virtual Vehicle
Research GmbH, Graz University of Technology, 8010 Graz, Austria.}
}

\maketitle
\thispagestyle{empty}
\pagestyle{empty}

\begin{abstract}

With the flourishing development of intelligent warehousing systems, the technology of Automated Guided Vehicle (AGV) has experienced rapid growth. Within intelligent warehousing environments, AGV is required to safely and rapidly plan an optimal path in complex and dynamic environments. Most research has studied deep reinforcement learning to address this challenge. However, in the environments with sparse extrinsic rewards, these algorithms often converge slowly, learn inefficiently or fail to reach the target. Random Network Distillation (RND), as an exploration enhancement, can effectively improve the performance of proximal policy optimization, especially enhancing the additional intrinsic rewards of the AGV agent which is in sparse reward environments. Moreover, most of the current research continues to use 2D grid mazes as experimental environments. These environments have insufficient complexity and limited action sets. To solve this limitation, we present simulation environments of AGV path planning with continuous actions and positions for AGVs, so that it can be close to realistic physical scenarios. Based on our experiments and comprehensive analysis of the proposed method, the results demonstrate that our proposed method enables AGV to more rapidly complete path planning tasks with continuous actions in our environments. A video of part of our experiments can be found at https://youtu.be/lwrY9YesGmw.

\end{abstract}

\section{INTRODUCTION}

With the development of the industrial digitalisation, intelligent warehousing systems \cite{c1} have become an important part of industrial production. Nowadays, Automated Guided Vehicle (AGV) \cite{c2} plays a crucial role in intelligent warehousing systems and its path planning has become the focus of research. The path planning algorithms \cite{c3,c4,c5} of AGVs develop rapidly. Researchers have proposed many classical path planning algorithms such as A* algorithm \cite{c6,c7}, Rapidly-Exploring Random Tree (RRT) \cite{c8}, Dynamic Window Approach \cite{c9} and Particle Swarm Optimization \cite{c10}, which have been widely used in simple environments. However, in realistic scenarios, most of these remain in simulation due to the computational complexity as well as the limitations of the real environment. 

At this time, Reinforcement Learning (RL) \cite{c11} has been studied to solve the path planning problem. NAIR et al. \cite{c13} proposed a path planning and obstacle avoidance method Modified Temporal Difference Learning for environment where static obstacles are known. On the basis of Temporal-Difference (TD) algorithm, WATKINS et al. \cite{c14} proposed Q-Learning algorithm that is widely used in discrete path planning environments. With the increasing complexity of the environments that agents need to process, Google's AI team DeepMind proposed the innovative concept of combining deep learning, which is the processing of perceptual signals, with RL to form Deep Reinforcement Learning (DRL) \cite{c15}. The DeepMind team proposed a new approach to DRL, Deep Q-Learning (DQN) \cite{c16}. With the successful application of DRL, much research has begun to explore DRL methods to solve problems of AGV path planning. Yang et al. \cite{c17} combined a priori knowledge and the DQN algorithm to solve the problem of slow convergence of AGVs in a warehouse environment. Panov et al. \cite{c18} studied the DQN algorithm to static grid maps and proved that the algorithm can obtain effective path planning. However, most of the related research on AGV path planning problems use 2D grid maps for experiments, which are still far from the actual physical environment. Real path planning environments are usually complex and thus current researches haven't solved the problem of slow searching of agents in sparse reward environments. In order to solve the above problems, we will propose a method that can improve intrinsic rewards and conduct experiments in simulation environments that approximate the real physical environment. 

As the Proximal Policy Optimisation (PPO) algorithm \cite{c19} has been proven to be widely applicable in complex environments, we use PPO as a deep reinforcement learning method in this paper. Xiao et al. \cite{c20} introduced distributed sample collection training policy and Beta policy for action sampling, which exhibits stronger robustness in the PPO algorithm. Our team \cite{c21} use a curiosity-driven model to enhance the exploration of the AGV agent. Shi et al. \cite{c22} studied a dynamic hybrid reward mechanism based on the PPO algorithm to solve the RL problem with sparse rewards. In DRL, reward shaping \cite{c23} can solve the problem of sparse reward environment, but constructing suitable reward functions is not easy, and in most cases, reward shaping limits the performance of algorithms such as PPO. In order to solve the sparse reward problem, we propose to introduce an exploratory method for the deep reinforcement learning algorithm PPO. The basic method of Random Network Distillation (RND) \cite{c24,c25,c26} is to increase the intrinsic rewards of agents and assist the extrinsic rewards to enable agents to better explore the environment. This has not been studied for path planning yet. Combining the PPO algorithm with the intrinsic reward measurements from RND, we augment the extrinsic reward in the environment during AGV path planning. In addition, in order to be able to simulate the path planning in real intelligent warehouses, we set up experimental environments for AGVs, and the experimental results show that by enhancing the extrinsic rewards through the intrinsic rewards of RND, our proposed method is able to explore several sparsely rewarded AGV path-planning environments more efficiently and stable. In summary, our contributions of this paper include the following two aspects.

\begin{itemize}
\item We propose a novel AGV path planning method RND-PPO, which combines the random network distillation mechanism with the PPO algorithm. Extrinsic rewards from environmental feedback are enhanced by additional intrinsic rewards, to solve the problem of AGVs that learn hard in sparsely rewarded path planning environments. 
\item We design simulation AGV agent path planning environments with physical rigid body properties and continuous motion space. The environments have both fixed and randomly generated target objects to simulate the real environment.
\end{itemize}

The rest of this paper is organized as follows. Section II describes the AGV path planning environment model. In Section III, the framework of our proposed RND-PPO method is presented and related algorithms are given in detail. The experiments and results are demonstrated in Section IV. Finally, Section V presents the conclusion and future work.

\section{AGV Path Planning Environment Model}
We design AGV path planning environment model with real physical body and action. For the situations that AGV agents need to face in real physical environments, we design multiple sets of models based on a simple scene and a complex scene, both of which consist of a closed square space, an AGV agent body, multiple static obstacles and a target object. The complex scene is four times the size of the simple scene. Most of the research is based on studies of static environments. In order to better test the performance of RND-PPO in different environments, we added dynamic target objects to these scenes. The simulation scenes are shown in Fig. 1. The simple scene on the left has two randomly generated targets and the complex scene on the right has three. Target objects are represented by red blobs and the agent is a blue blob.

The AGV agent is described by a set of state variables and interacts with the environment by performing actions to change its state variables. To replace the discrete actions used in most research, we build the agent as rigid body and create a continuous action space for it. The continuous actions are generated from a neural network and then passed to the action function, which processes the received action vectors. In this paper, the environment contains two consecutive vectors representing the control forces in the $X$-axis and $Z$-axis, which are transmitted to the physical force to make agent move.

During training, the sensors provide state information to the agent, such as the position, velocity, colour of other objects in the environment and so on. In our model, the internal observation space dimension of the agent is 8, which records the 3D position of agent, the 3D position of target object, and the $X$-axis component and $Z$-axis component of the speed of the agent are observed respectively. In addition, the AGV agent is equipped with two 3D ray perception sensors, one for detecting the information around the agent with 10 rays and horizontal field of view of 360 degrees. The other one intensively detects information in the forward direction of agent, with 7 rays and horizontal field of view of 120 degrees. Each ray can detect 2 targets including wall and target object, and each ray has two dimensions to detect collision or not, so the total observation dimensions are $(10+7) \times (2+2)$.

\begin{figure}[t]
    \centering
    \includegraphics[width=1\linewidth]{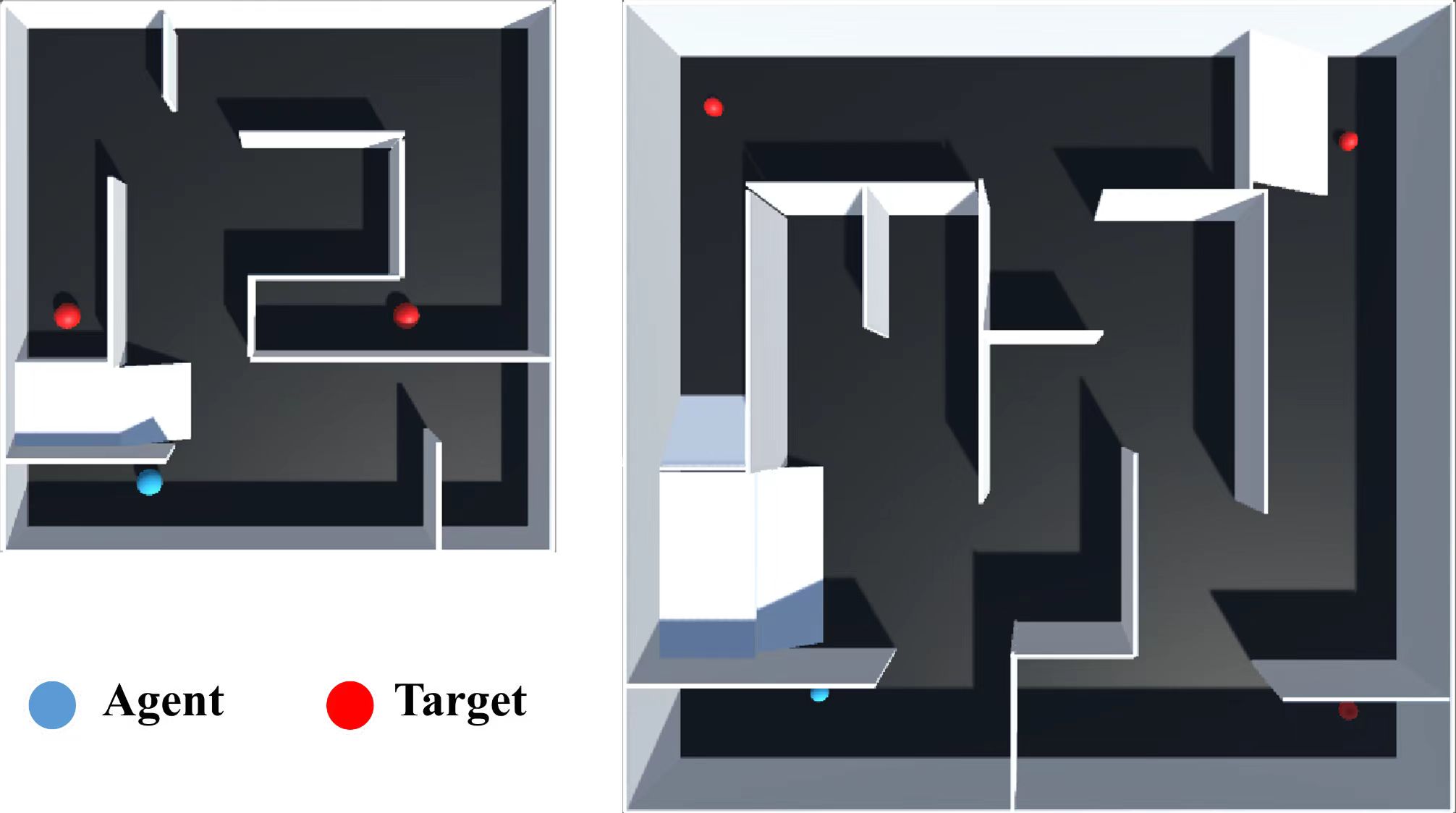}
    \caption{Top view schematic of AGV path planning simulation scenario.}
    \label{fig:1}
\end{figure}

The computation of reward function within each learning episode is divided into two parts. One for the extrinsic reward $r_e$ obtained from the interaction of agent with environment, and the other for the intrinsic reward $r_i$ given to agent by our proposed RND-PPO model, which will be introduced in the next section. Thus the total reward is written as
\begin{equation}
        r = r_t^e + r_t^i.
\end{equation}
When the agent collids with the target object, the extrinsic reward is set to 5. This value is an empirical value obtained from several experiments. In order to make the agent find the target as soon as possible, we design a tiny single-step negative reward. The extrinsic reward function is defined as 
\begin{equation} 
	r_t^e = \begin{cases}
                \frac{-1}{\text{MaxStep}} & \text{single-step} \\
                5 & \text{agent collides target object} \\
                \end{cases}
    \end{equation}
The MaxStep is the maximum number of steps an agent can explore in a learning episode.

\section{AGV PATH PLANNING BASED ON RND-PPO}

It is often impractical to design dense reward functions for tasks of RL agents, so agents need to explore the environment in a targeted manner. RND \cite{c24} was introduced as an exploration method for DRL methods, and it has the flexibility to combine intrinsic and extrinsic rewards.

\begin{figure}[t]
    \centering
    \includegraphics[width=1\linewidth]{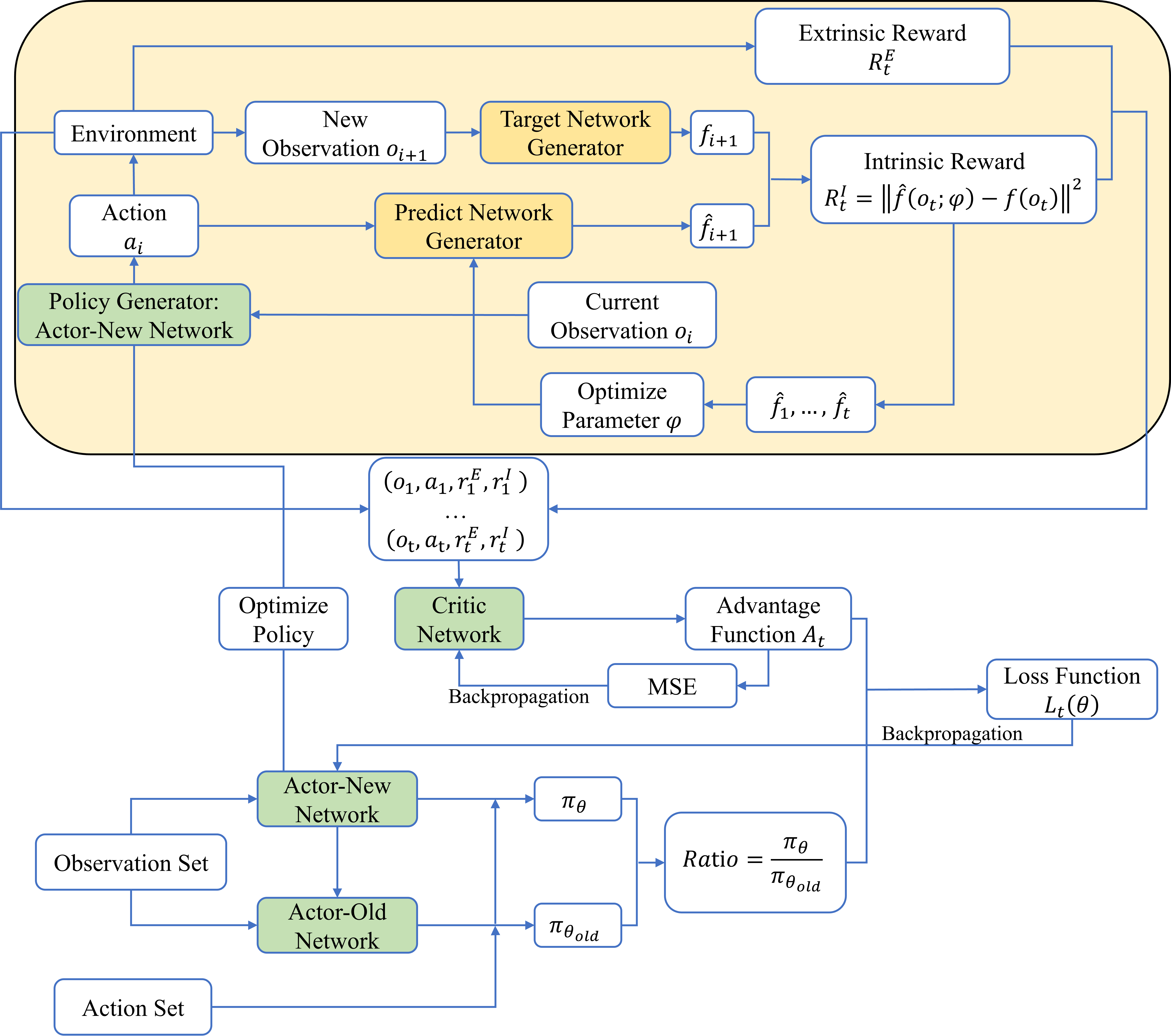}
    \caption{Structure of the proposed RND-PPO.}
    \label{fig:2}
\end{figure}

\subsection{Framework of AGV Path Planning with RND-PPO}
The key to solving the AGV path planning problem using RL is how the AGV agent updates its own action policy based on the received rewards to obtain the maximum cumulative reward value. In the real AGV path planning environments, the rewards are sparse, in this case we need to use intrinsic rewards to guide the AGV agent to fully explore the state space and action space in the environment, so we design a new exploration mechanism RND-PPO to motivate the agent to explore the environment. 

The structure of our proposed RND-PPO is shown in Fig. 2. The AGV agent training process is divided into two stages. The yellow box is the RND stage and the other part connected to it is PPO training stage. RND defines a new training stage and the RND training alternates with the training of the agent. The model obtained from the RND training is input to PPO and used to generate the corresponding intrinsic rewards. The next stage is the agent training stage, which is a stage of using the trained RND model, combining the intrinsic rewards predicted by the RND model with the RL algorithm PPO. In the end, the agent completes the learning of the optimal policy by using the obtained extrinsic rewards $r_t^e$, intrinsic rewards $r_t^i$ and environment state $s_t$.

\subsection{Random Network Distillation Model} 

\begin{algorithm}[!b]
    \caption{RND-PPO Pseudocode}

   \begin{algorithmic}[1] 
        \STATE Input: training epochs per collect $E$; batch size $B$; number of learning episodes $M$; length of learning episodes $N$;  number of predict optimization steps $N_{pre}$
        \STATE Initialize policy network parameters $\theta$
        \STATE Initialize fixed target network parameters $\varphi$ and prediction network parameters $\hat{\varphi}$
    
        \FOR{$m = 1$ to $M$} 
            \STATE Collect a set of trajectories $T_m= \{s_t,s_{t+1},a_t,r_t^e\}$ by run policy $\pi_{\theta_m}$ $m$ Timesteps
            \STATE Update observation normalization parameters by $T_m$
            \FOR{$i = 1$ to $N_{pre}$}
                \STATE Sample $a_t \sim \pi(a_t | s_t)$
                \STATE Sample $s_{t+1}, r_t^e \sim p(s_{t+1}, r_t^e|s_t,a_t)$ 
                \STATE Calculate intrinsic reward $r_t^i$ = $\big\| \hat{f}(s_{t+1} ; \hat{\varphi}) - f(s_{t+1};\varphi) \big\| ^2$
                \STATE Optimize $\hat{\varphi}$ w.r.t. distillation loss $r_t^i$ using Adam
            \ENDFOR
            \STATE Normalize $r_t^i$, obtain normalized intrinsic reward $\hat{r}_t^i$
            \STATE Normalize $r_t^e$, obtain normalized extrinsic reward $\hat{r}_t^e$
            \STATE Calculate total reward $r_t = \alpha \hat{r}_t^e + \beta \hat{r}_t^i$, and obtain the final trajectories $\hat{T}_m = \{s_t,s_{t+1},a_t,r_t\}$
            \STATE Calculate advantage estimates $\hat{A}_{\theta_m}$ using Eq. (4) with value on $\hat{T}_m$
            \FOR{$e=1$ to $E$}
                \STATE Sample minibatch $b$ episodes from $\hat{T}_m$
                \STATE Update policy parameters $\theta$ by maximizing $L(\theta)$ in Eq. (6) with Adam, where ratio is used by Eq. (5)
            \ENDFOR
        \ENDFOR
    \end{algorithmic}
\end{algorithm}
To address the lack of exploration of PPO in sparse reward environments, among the intrinsic reward methods used for exploration, we invoke Random Network Distillation (RND) which is a technique based on prediction error. The model is presented in the yellow part of Fig. 2. In RND, the agent first constructs a randomly-fixed target neural network $f$, where fixed means that it will not be updated throughout the learning process, and constructs a prediction network $\hat{f}$, whose goal is to predict the output of the randomly-set target network $f$. The target network $f$ and the prediction network $\hat{f}$ map the observations $\mathbb{S}$ to the reward $\mathbb{R}^k$. The target network defined as: $f: \mathbb{S} \rightarrow \mathbb{R}^k $, network parameter is denoted as $\varphi$ and remain fixed after random initialisation. The prediction network defined as: $\hat{f}: \mathbb{S} \rightarrow \mathbb{R}^k $, network parameter is denoted as $\hat{\varphi}$ which is trained to minimise the prediction error. The parameter $\hat{\varphi}$ is updated by minimising the expected value of the mean square error $\big\| \hat{f}(s_t ; \hat{\varphi}) - f(s_t;\varphi) \big\|^2$ through gradient descent algorithm. The agent will input the observation $s_t$ obtained from the environment into the target network $f$, at which time $f(s_t)$ serves as the prediction target of the prediction network $\hat{f}$. When the prediction network $\hat{f}(s_t)$ is input with a novel state, due to the large discrepancy between this distribution and inputs it has ever received, the agent will receive a large intrinsic reward as 
\begin{equation}
        r_t^i = \big\| \hat{f}(s_t ; \hat{\varphi}) - f(s_t;\varphi) \big\| ^2.
\end{equation}

\begin{figure}[!b]
    \centering
    \includegraphics[width=0.8\linewidth]{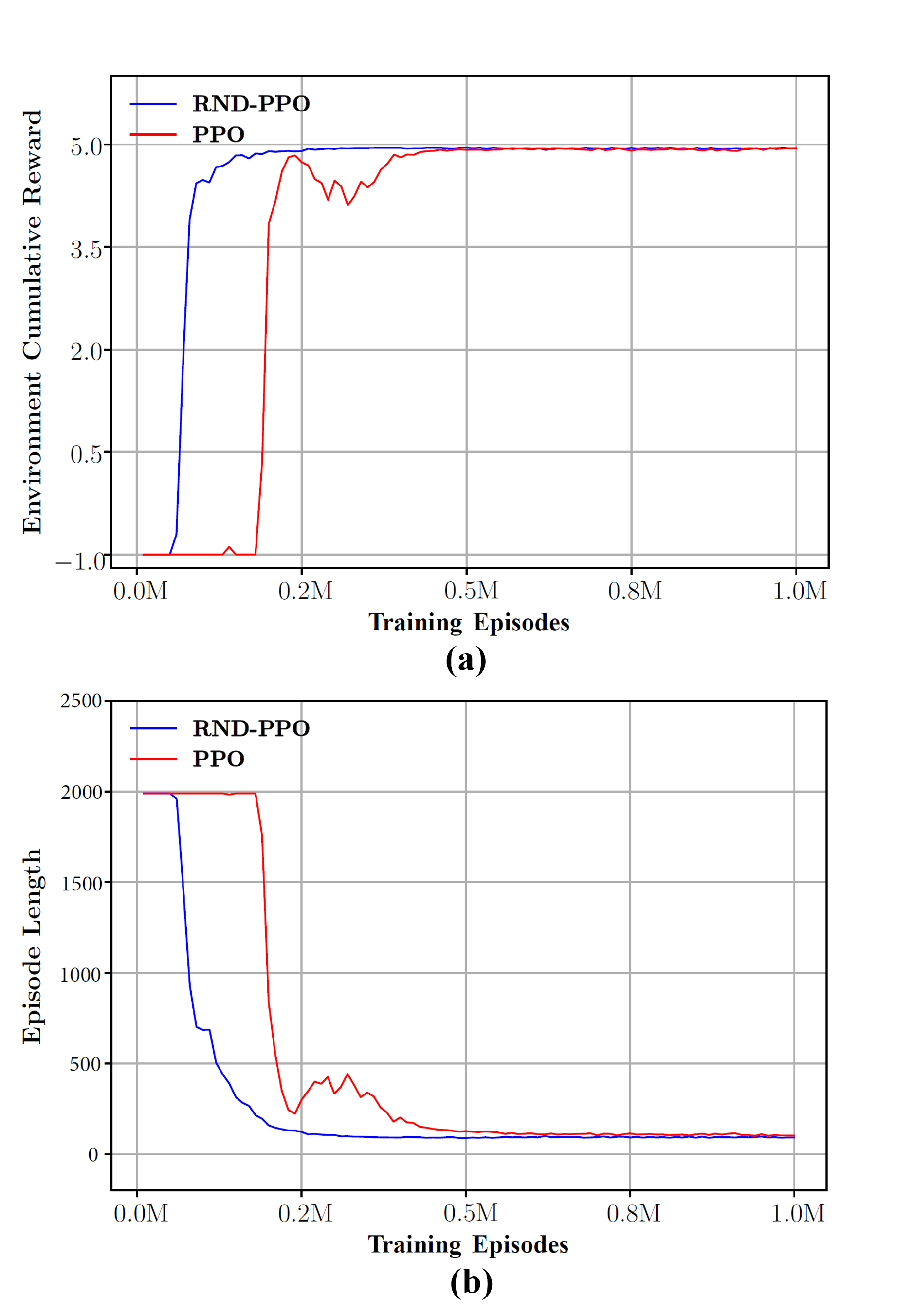}
    \caption{Behabior of reward and episode length in the simple static scenario. (a) environment cumulative reward of the AGV agent and (b) episode length of the AGV agent.}
    \label{fig:3}
\end{figure}
\subsection{AGV agent path planning with RND-PPO}
In the sparse reward environment, we propose an exploration mechanism that uses RND based PPO to motivate the agent to find more novel state $s$. First, we give the concept of state novelty which can be measured by the prediction error. For AGV agent observing the state $s$ at the current moment, the fewer the number of states similar to state $s$ among all previously visited states, the more novel state $s$ is.

The PPO algorithm is essentially a model-free algorithm, and its core architecture remains an Actor-Critic algorithm. The critic network fits the state value function and action value function through the environmental state information $s$ observed by an agent, and updates critic network parameters by calculating advantage function and using the mean square error as critic loss function. The advantage function is shown as
\begin{equation}
        A_t = \sum_{t^\prime > t} \beta^{t^\prime-t} r_{t^\prime} - V_\pi(s;\theta),
\end{equation}
where $\beta$ is a tunable coefficient. Different estimates can be obtained by adjusting $\beta$. When updating the actor network, PPO uses two networks with the same structure to preserve the old and new policy. The policy ratio is used to measure the ratio of the probability of taking a certain state-action pair $(s,a)$ under the new policy to the probability of taking the same state-action pair under the old policy. The policy ratio is defined as
\begin{equation}
        r_t(\theta) = \frac{\pi_\theta(a_t | s_t)}{\pi_{\theta_{old}(a_t | s_t)}}.
\end{equation}

PPO introduces a new clip mechanism, which can effectively reduce the number of computation steps while limiting the magnitude of policy update, and it is defined as follows
\begin{equation}
        L(\theta) = \mathbb{E}_t [\text{min}(r_t(\theta) \hat{A_t},\text{clip}(r_t(\theta),1-\epsilon,1+\epsilon)\hat{A_t})],
\end{equation}
where $\epsilon$ is a hyperparameter, $\hat{A_t}$ is an estimate of the advantage function at time step $t$. The purpose of setting $1-\epsilon,1+\epsilon$ is to specify the magnitude of the policy update to prevent the update from being too large and causing the training to be unsmooth.
\begin{figure}[b!]
    \centering
    \includegraphics[width=0.8\linewidth]{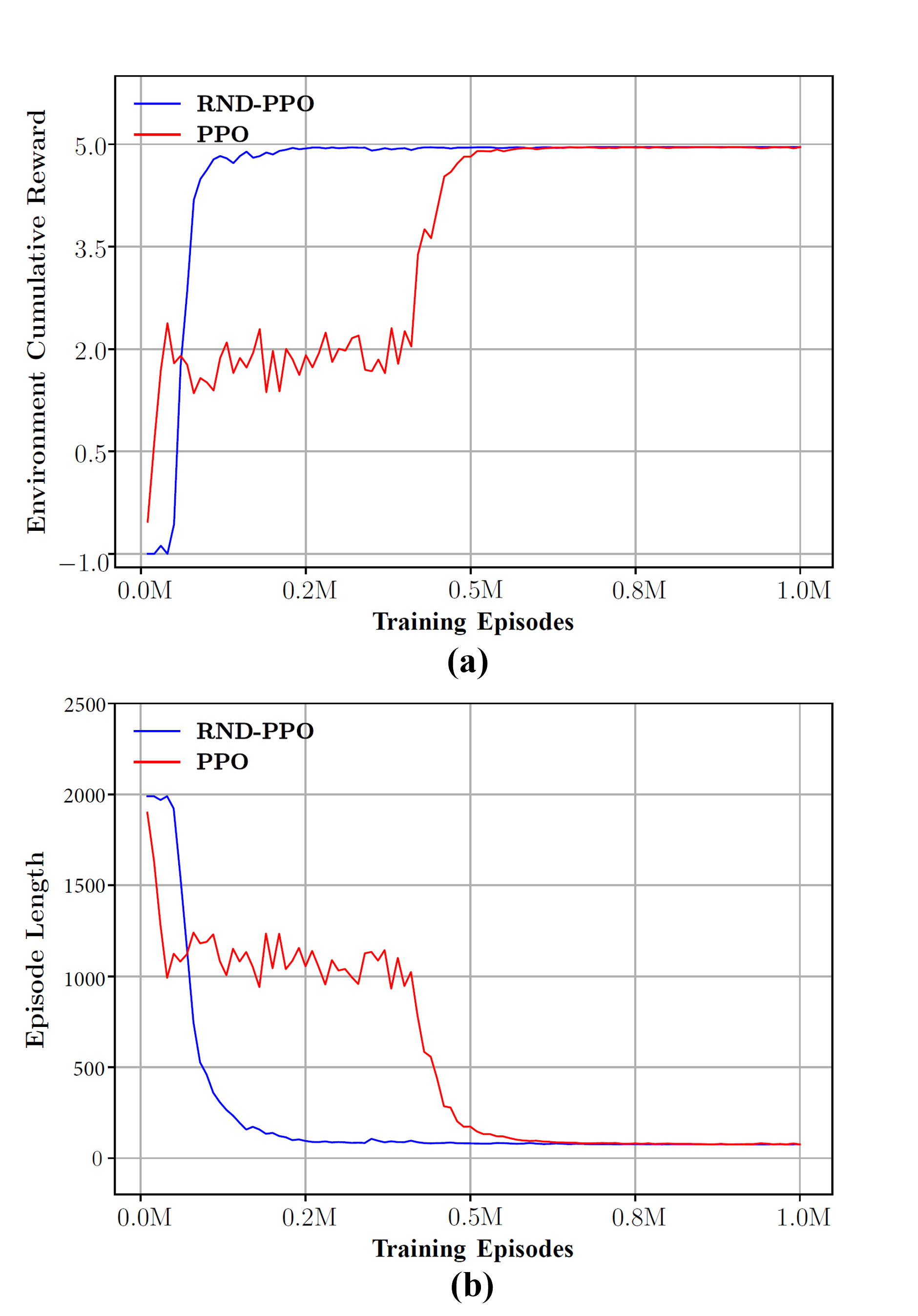}
    \caption{Behavior of reward and episode length in the simple dynamic scenario. (a) environment cumulative reward of the AGV agent and (b) episode length of the AGV agent.}
    \label{fig:4}
\end{figure}

In our proposed method shown in Algorithm 1, the first three lines initialise various parameters of the RND-PPO. After that, the training process of the RND model starts to indicate lines in Algorithm 1. The parameters of the target network $\varphi$ are fixed, and according to the stochastic gradient descent method, the expected value of the mean square error $\big\| \hat{f}(s_t ; \hat{\varphi}) - f(s_t;\varphi) \big\|^2$  is minimised, and prediction network parameters $\hat{\varphi}$ are optimised. This RND process can be regarded as doing distillation between the target network, which is randomly generated with fixed parameters, and the prediction network, whose parameters are to be updated, so that the prediction network is constantly close to the target network. Then the training process of the agent using the PPO algorithm based on intrinsic and extrinsic rewards starts to indicate lines in Algorithm 1. These rewards are first normalised separately to compute the final set of training trajectories. In the last stages, it combines intrinsic motivations with extrinsic rewards to calculate the advantage function and value function, subsequently refining PPO by updating the policy parameters $\theta$.

\section{Experiments}
In this section, we evaluate our method RND-PPO, for learning AGV agent path planning policy in two groups of experiments. First, we introduce details of our implementation, including the hyperparameters. Then, we compare our proposed method with the baseline PPO in both static and dynamic scenarios. Static and dynamic environments also include simple and complex scenarios, respectively. Experiments show that using RND can improve the efficiency and stability of AGV agent learning path planning policy.

\begin{figure}[!t]
    \centering
    \includegraphics[width=1\linewidth]{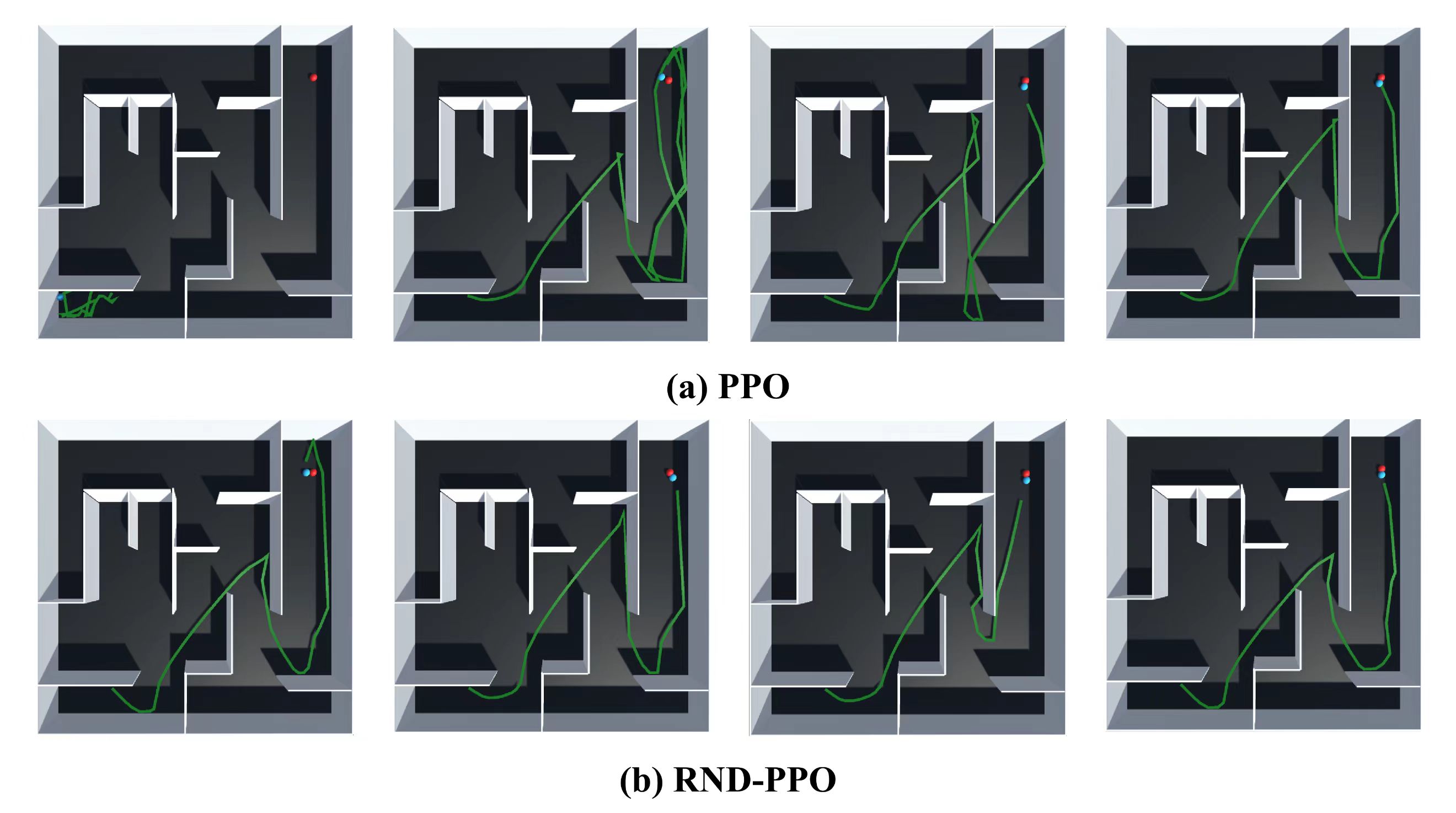}
    \caption{Complex static environment path planning trajectories. From left to right, the training episodes are $0.25$, $0.5$, $0.75$ and $1.0\cdot 10^6$. (a) corresponds to the PPO and (b) corresponds to the RND-PPO.}
    \label{fig:5}
\end{figure}

\subsection{Experimental Settings}
The AGV agent body and the target object are spheres with a radiuis of $0.5$ per unit length, the size of the simple scene is $20\times20$, the size of the complex scene is $40\times40$. The maximum number of steps for each learning episode of the static and dynamic experiments in the simple scene is 2000, and the maximum number of steps for the complex static and dynamic scenes is 3000 and 4000, respectively. The number of learning episodes in each experiment is $1\cdot 10^6$. The reward function of AGV agent is shown in Eq. (1). All the experiments are carried out on an AMD Ryzen 7 5800H 3.20 GHz PC with 16GB memory.

\subsection{Simple Scene Experiments}

The simple scene is a $20\times20$ map: $(-5.0,0.5,-8.0)$ is the start location of agent, and $(5.0,0.5,-1.5)$ is the location of target object in static experiment. In dynamic experiment, the target object will be randomly generated in $(5.0,0.5,-1.5)$ and $(-8.0,0.5,-1.0)$.

We test our method in the simple scene and choose three metrics including training episodes, environment cumulative reward and episode length, to evaluate our experimental results. First, we test our proposed method in the simple static environment. In the simple static environment, after $0.18\cdot 10^6$ episodes of training, our proposed method RND-PPO is able to obtain the environmental cumulative reward of $4.8$ within $280$ steps of an episode. In contrast, as shown in Fig. 3, the PPO algorithm without RND performs poorly, and the agent is able to obtain the same environmental reward value within $238$ steps over $0.39\cdot 10^6$ episodes of training. However, performance of PPO is worse in the simple dynamic environment where there are two randomly generated target objects, and since there is no intrinsic reward for exploration of the environment. It is difficult for PPO to explore the location of the other target object that would earn a reward as shown in Fig. 4. Although PPO relies on search by chance to find the first target object faster than RND-PPO, it needs to spend a large number of training episodes in searching the second target object. PPO can find the target object with an average of $170$ steps after requiring $0.52\cdot 10^6$ training episodes and get an environmental cumulative reward of $4.8$. In contrast, our proposed RND-PPO method only requires $0.18\cdot 10^6$ training episodes to get same environmental cumulative reward with an average of $187$ steps. 

\begin{figure}[!t]
    \centering
    \includegraphics[width=1\linewidth]{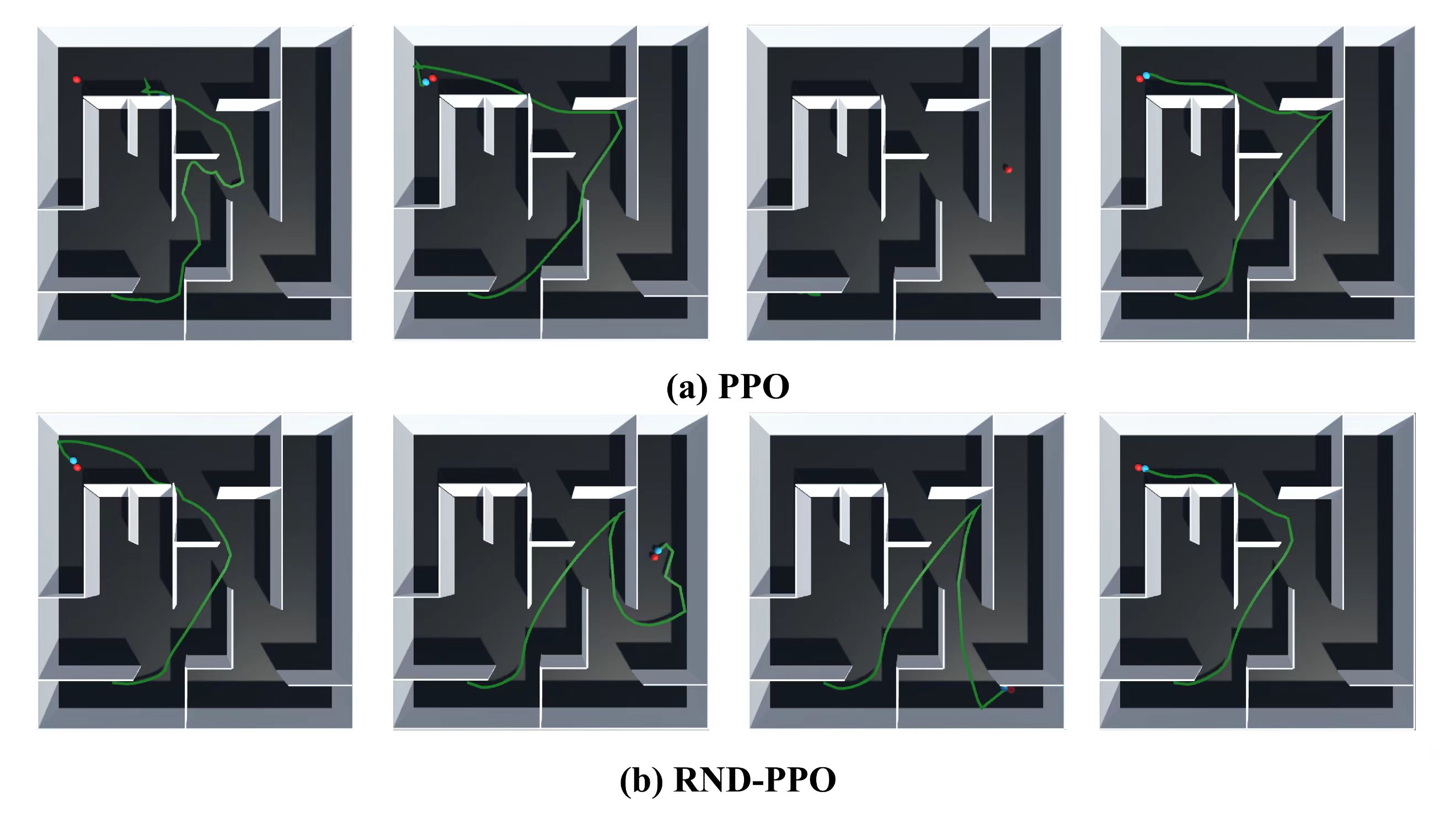}
    \caption{Complex dynamic environment path planning trajectories. From left to right, the training episodes are $0.25$, $0.5$, $0.75$ and $1.0\cdot 10^6$. (a) corresponds to the PPO and (b) corresponds to the RND-PPO.}
    \label{fig:6}
\end{figure}

\subsection{Complex Scene Experiments}

The complex scene is a $40\times40$ map: $(-12.0,0.5,-16.0)$ is the start location of agent, and $(17.0,0.5,15.0)$ is the location of target object in the static experiment. In the dynamic experiment, the target object will be randomly generated in $(15.0,0.5,2.0)$, $(15.0,0.5,-17.0)$ and $(-17.0,0.5,15.0)$. Fig. 5 and Fig. 6 show the trajectory of the AGV agent after training $0.25$, $0.5$, $0.75$ and $1.0\cdot 10^6$ episodes in the complex static and dynamic environment respectively. The AGV agent trained using RND-PPO has found the path to reach the target object after $0.2\cdot 10^6$ episodes, while the agent trained only by PPO is still exploring the space around the starting position. The two metrics used for evaluation can be found in Fig. 7, and our proposed method is the better performer on both data. In the static environment, after the same training of $0.2\cdot 10^6$ episodes, our proposed method is already able to obtain an environmental cumulative reward of $4.85$ in $492$ steps of an episode, while the PPO can hardly get an environmental reward value of $-1$, which means it still expores the environment. The agent trained by PPO is able to reach the same environmental reward value of $4.85$ only after beeing trained for at least $0.34\cdot 10^6$ episodes.

In the dynamic experiment, Fig. 6 shows the experimental results of the AGV agent after being trained by PPO and RND-PPO. It can be seen that our proposed method RND-PPO can find the optimal path quickly and accurately when the target object randomly appears in three positions. However, the agent trained only by using PPO can only find the target object located in the upper left corner due to the fact that there is no intrinsic reward that can motivate the AGV agent to explore the whole environment. The relationship between the three metrics is shown in Fig. 8. The agent trained by RND-PPO found the first target object after $0.07\cdot 10^6$ episodes of training, while PPO did not complete this goal until around $0.16\cdot 10^6$ episodes. During the $0.08-0.16\cdot 10^6$ episodes of training, the curve of RND-PPO fluctuated due to the presence of dynamic objects, and fell into a short struggle in exploring the new environment. But soon with the help of the intrinsic rewards of RND, the agent learnt the paths to reach the three target objects. The agent under RND-PPO training is able to reach more than 4.8 environment cumulative reward after $0.24\cdot 10^6$ episodes with $257$ steps per episode. The PPO, on the other hand, still failed to complete the entire path planning task until the end of training. In conclusion, our proposed method explore static and dynamic environment faster in both simple or complex scene than the agent trained with PPO only.

\begin{figure}[t]
    \centering
    \includegraphics[width=0.8\linewidth]{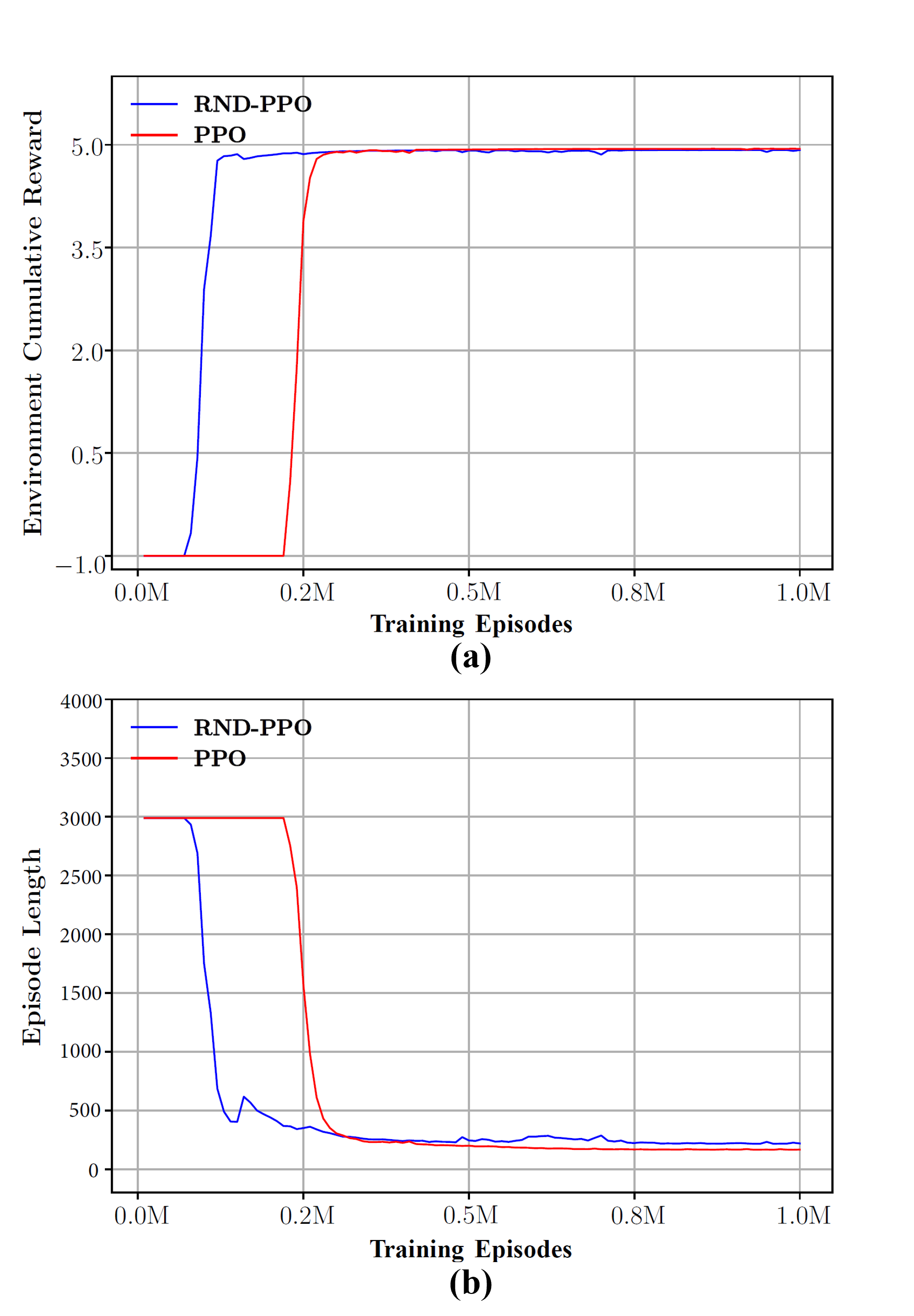}
    \caption{Behavior of reward and episode length in the complex static scenario. (a) environment cumulative reward of the AGV agent and (b) episode length of the AGV agent.}
    \label{fig:7}
\end{figure}

\section{CONCLUSIONS}

In this paper, we propose a novel method RND-PPO for AGV path planning, which introduces random network distillation mechanism to give intrinsic rewards to the AGV agent to address the effect of sparse reward environments and to improve the speed of training. In addition, we have developed simulated environments with realistic physical states containing the location of static obstacles and dynamic targets. We evaluate our approach with different scenarios. Both qualitative and quantitative experiments show that our approach is efficient with good performance. The RND-PPO agent makes use of intrinsic rewards, avoids limiting itself to a single rewarded target object, and adapts quickly to changes in the external environment. We adopt the widely used PPO algorithm as the basic implementation, which can in principle be extended to other RL algorithms (e.g., SAC). Our future work will focus on statistical analysis of RND-PPO in more complex dynamic environments to optimise the use of intrinsic rewards.

\begin{figure}[t]
    \centering
    \includegraphics[width=0.8\linewidth]{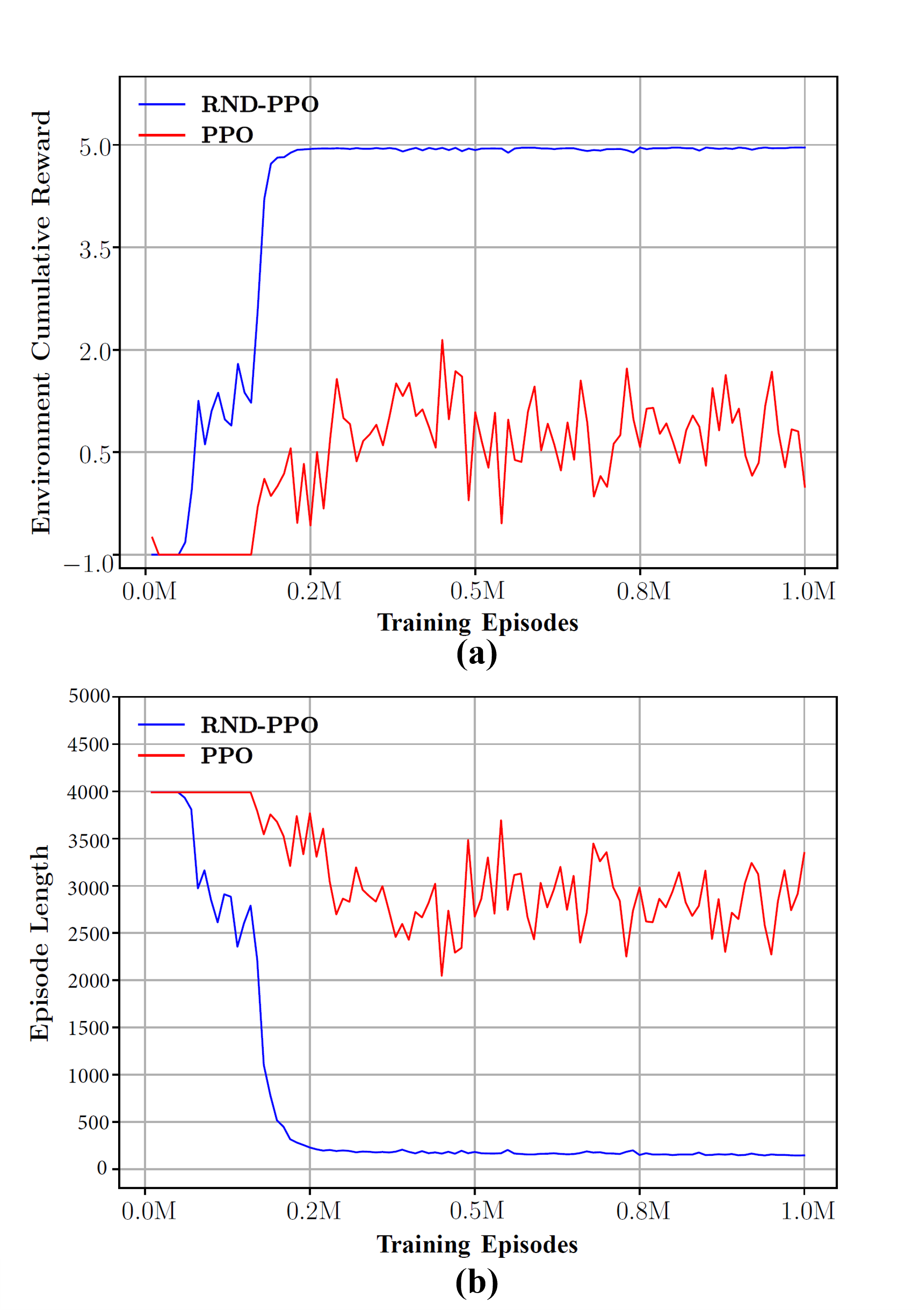}
    \caption{Behavior of reward and episode length in the complex dynamic scenario. (a) environment cumulative reward of the AGV agent and (b) episode length of the AGV agent.}
    \label{fig:8}
\end{figure}

\addtolength{\textheight}{-12cm}   




\section*{ACKNOWLEDGMENT}

This work was supported by the National Natural Science
    Foundation of China under Grant No. 62133011 and the Special Funds of the Tongji University for "Sino-German Cooperation 2.0 Strategy" No. ZD2023001. 
    The authors would like to thank TÜV SÜD for the kind and
    generous support. We are also grateful for the efforts from our
    colleagues in Sino German Center of Intelligent Systems in Tongji University.


\end{document}